\documentclass[10pt,twocolumn, letterpaper]{article}
\usepackage{cvpr}
\usepackage{orcidlink} 

\usepackage[font=footnotesize]{subcaption}

\usepackage[normalem]{ulem}
\usepackage{tabularray}
\UseTblrLibrary{booktabs}

\usepackage{hyperref}
\title{P-TAME: Explain Any Image Classifier with Trained Perturbations\thanks{This work was supported by EU Horizon Europe Program under Grant 101070190 AI4TRUST. }}
\author{Mariano V. Ntrougkas\orcidlink{0009-0004-0569-0837}$^{1,2}$, Vasileios Mezaris\orcidlink{0000-0002-0121-4364}$^{1}$, and Ioannis Patras\orcidlink{0000-0003-3913-4738}$^{2}$\\
$^{1}$Information Technologies Institute / CERTH, Thessaloniki, Greece\\
$^{2}$Queen Mary University of London, London, UK\\
Corresponding author: Mariano V. Ntrougkas (email: ntrougkas@iti.gr).}
\begin{document}
\maketitle
\begin{abstract}
The adoption of Deep Neural Networks (DNNs) in critical fields where predictions need to be accompanied by justifications is hindered by their inherent black-box nature. This paper introduces P-TAME (Perturbation-based Trainable Attention Mechanism for Explanations), a model-agnostic method for explaining DNN-based image classifiers. P-TAME employs an auxiliary image classifier to extract features from the input image, bypassing the need to tailor the explanation method to the internal architecture of the backbone classifier being explained. Unlike traditional perturbation-based methods, which have high computational requirements, P-TAME offers an efficient alternative by generating high-resolution explanations in a single forward pass during inference. We apply P-TAME to explain the decisions of VGG-16, ResNet-50, and ViT-B-16, three distinct and widely used image classifiers. Quantitative and qualitative results show that P-TAME matches or outperforms previous explainability methods, including model-specific ones.\footnote{Code and trained models are available at \url{https://github.com/IDT-ITI/P-TAME}.}
\end{abstract}
\section{Introduction}
\label{sec:intro}

Advances in deep neural networks (DNNs) over the past decade have been tremendous. However, a persistent challenge is the lack of DNN explainability \cite{liInterpretable2022}. DNNs are often referred to as ``black-box'' models since they do not provide users with insights into their decision-making process, impeding their wider adoption in important application domains such as healthcare, journalism, and law enforcement, where the ability to justify decisions is a critical requirement \cite{salahuddinTransparency2022, Pavlidis02012024}. 
\begin{figure}[t]
  \centering
  \includegraphics[width=\linewidth]{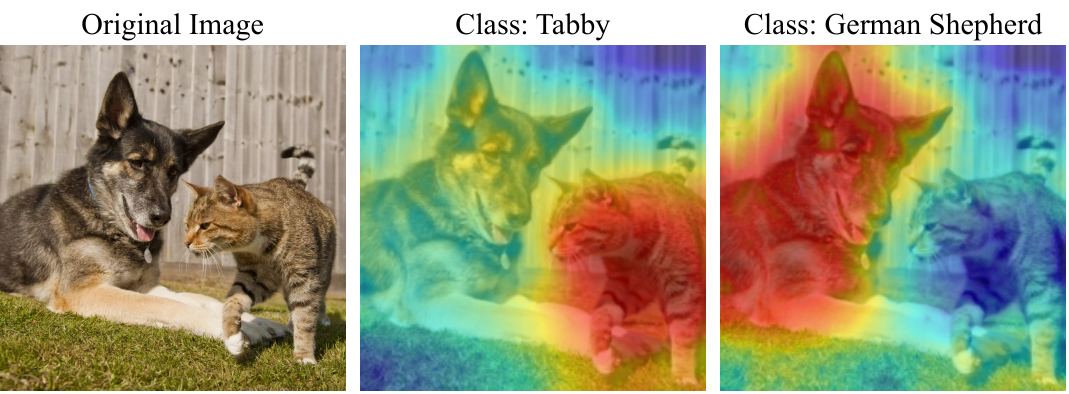}

  \caption{Example class-specific explanations produced by the P-TAME method.
  The image classifier whose predictions are being explained is the ViT-B-16 model, using the default weights from torchvision.}
  \label{fig:example}
\end{figure}
Consequently, there is a growing interest in developing methods to make DNN decisions more understandable to users, i.e., in developing eXplainable Artificial Intelligence (XAI) methods \cite{liInterpretable2022}.

Within this research domain, a dominant direction to advancing the explainability of DNN image classifiers is to generate saliency maps \cite{saleemExplaining2022}, which highlight the regions of the input image that are most relevant to the decision of the DNN. Saliency maps (a.k.a. explanation maps; Fig.~\ref{fig:example}) can help users
understand why a DNN made a particular decision and can also be used to identify potential biases in the decision-making process \cite{lapuschkinUnmasking2019}.
\begin{figure*}[t]
  \centering
  \includegraphics[width=0.85\linewidth]{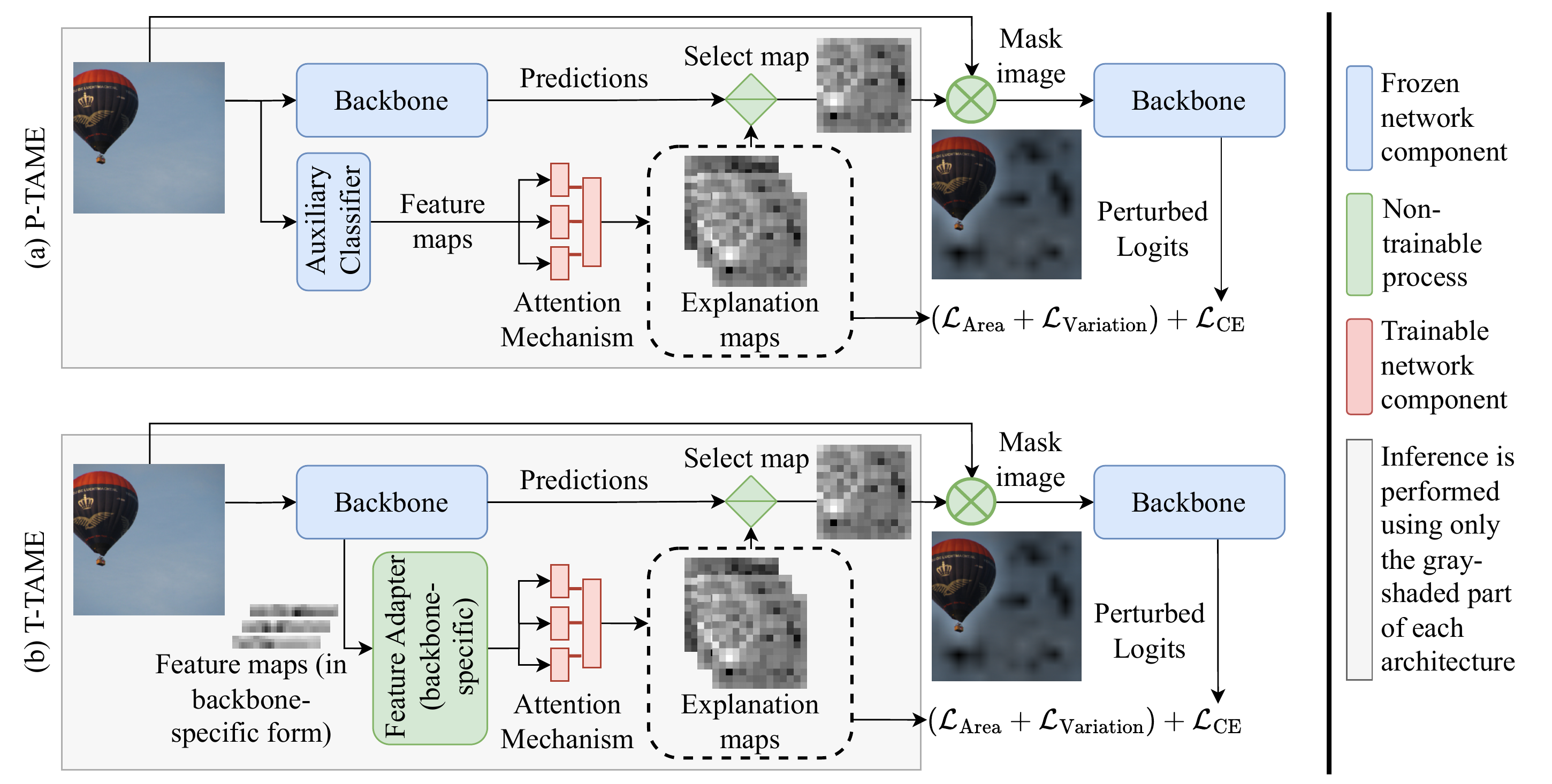}

  \caption{Overview of the proposed P-TAME method (a),  displaying the pipeline for both the training and inference stages. In this illustration, the main difference between the P-TAME method (a) and T-TAME (b) is evident: In P-TAME, no intermediate feature maps are extracted from the backbone (i.e., the DNN classifier whose decisions we aim to explain). Instead, an auxiliary classifier is employed to extract feature maps from the input image.}
  \label{fig:pipeline}
\end{figure*}

Several classes of methods have been proposed to generate saliency maps for DNNs, including gradient-based \cite{selvarajuGradCAM2017, chattopadhayGradCAM2018}, perturbation-based \cite{wangScoreCAM2020, petsiukRISE2018, desaiAblationCAM2020}, and response-based \cite{sattarzadehExplaining2021, sudhakarAdaSise2021, zhouLearning2016, ntrougkasTAME2022, ntrougkasTTAME2024} methods.
Gradient-based methods compute the gradient of the output with respect to the input image and use it to generate the saliency map. They suffer from the vanishing gradient problem and can be noisy and unreliable \cite{adebayoSanity2018}. Additionally, the most widely used methods in this category, Grad-CAM and Grad-CAM++ \cite{selvarajuGradCAM2017, chattopadhayGradCAM2018} require the extraction of intermediate feature maps from the network being explained, thus needing to be adapted to the DNN architecture of interest.
Perturbation-based methods generate saliency maps by perturbing the input image and observing the change in the output. They are more robust and reliable than gradient-based methods but are computationally expensive at the inference stage, which is a critical constraint in e.g., real-time edge applications in industrial inspection, where explanations must be generated in a resource-constrained environment \cite{hassanien2023explainable}. Furthermore, they are not always model-agnostic; e.g., the widely used Score-CAM \cite{wangScoreCAM2020} relies on extracting intermediate feature maps (similarly to Grad-CAM, Grad-CAM++).
Response-based methods, e.g., CAM \cite{zhouLearning2016}, generate saliency maps by combining the intermediate feature maps of the DNN to predict the saliency map.
They are, by definition, model-specific and often also make use of perturbations, making them computationally expensive. 

To address the limitations of previous methods, i.e., the noisy explanations produced by gradient-based methods and the computationally expensive process of using perturbations, TAME \cite{ntrougkasTAME2022} and T-TAME \cite{ntrougkasTTAME2024} (Transformer-compatible Trainable Attention Mechanism for Explanations) proposed a new paradigm for explaining the decisions of DNNs by generating saliency maps with an attention mechanism. 
The attention mechanism learns to combine the intermediate feature maps from multiple layers of the DNN to predict the saliency map. The quality of the produced explanation maps is generally on par with perturbation-based methods, while avoiding the need for multiple forward passes during inference. A limitation of many previous methods that persists, however, is that feature maps need to be extracted from the DNN that is being explained. Additionally, depending on the DNN architecture, these feature maps may need to be adapted to the T-TAME attention mechanism. Thus, TAME and T-TAME are not model-agnostic, in contrast to many perturbation-based approaches.

In this work, we propose P-TAME (Perturbation-based Trainable Attention Mechanism for Explanations), an attention-based XAI method that generates saliency maps directly from the input images using an auxiliary classifier, without the need to extract and process intermediate feature maps from the DNN being explained.
Since P-TAME is model-agnostic, it can be applied to any DNN image classifier. It produces saliency maps by learning to perturb the input image to highlight the regions most relevant to the decision of the DNN being explained; and, after training, produces explanations in a single forward step. 
The performance of P-TAME is evaluated, both quantitatively \cite{chattopadhayGradCAM2018,rongConsistent2022} and qualitatively, on three popular image classifiers: VGG-16 \cite{simonyanVery2015}, ResNet-50 \cite{heDeep2016}, and ViT-B-16 \cite{dosovitskiyImage2020} trained on ImageNet \cite{russakovskyImageNet2015}. 
Experimental comparisons demonstrate that P-TAME rivals state-of-the-art (SoA) perturbation methods in explanation quality, without needing multiple forward passes during inference. We provide P-TAME as an open-source library to support adoption and further XAI research.

In summary, the contributions of this work are as follows:
\begin{itemize}
  \item We propose P-TAME, a method that employs an auxiliary classifier to extract feature maps, which are then processed by an attention mechanism to generate explanation maps. P-TAME is model-agnostic, thus can be easily applied to any DNN image classifier.
  \item We evaluate P-TAME quantitatively on three popular image classifiers with very different architectures trained on the ImageNet dataset, and we compare it against T-TAME and other SoA methods.
\end{itemize}

\section{Related work}
\label{sec:related}

Humans have long explained and justified their actions, a core aspect of how they relate, cooperate, and build trust \cite{cappelenMaking2021}. Conversely, the inability of current AI systems to provide justifications for their actions hampers trust in their decisions.
There are many ways to increase trust and transparency of AI systems, but in this work we will focus on techniques that produce explanations for the decisions of image classifiers. The form of these explanations varies, depending on the nature of the data which the AI system is designed to work on.
For image classifiers, the most common form for explanations is a feature attribution map, also known as explanation map (Fig.~\ref{fig:example}). These explanations are local, as opposed to global, because they explain a single decision of the classifier, instead of describing how an image classifier reaches its decisions in general. In addition to explanation maps, recently, methods that aim to explain a classifier's decisions using a set of semantic concepts have been introduced, e.g., \cite{ghorbaniAutomatic2019, sunExplain2023, kumarMACE2021}. However, due to the considerably different form of the produced explanations, the latter methods are not comparable to XAI methods producing explanation maps.

This section establishes a brief taxonomy of XAI methods (for a more comprehensive taxonomy, refer to \cite{schwalbecomprehensive2024}) and describes notable XAI methods for image classifiers.
To produce explanations for an image classifier, we can employ an intrinsically explainable AI model (e.g. \cite{cosciaFeatures2024}), called an ante-hoc explainable model, or apply an XAI method to a trained model without modifying it. 
The latter methods are called post-hoc and have the advantage of being applicable to SoA
image classifiers without trading off performance for explainability. 
Post-hoc methods are further divided into model-agnostic and model-specific methods. Model-agnostic methods only require access to the model input and output, while model-specific methods require access to the model architecture and may have specific requirements to be applicable. 
A relative drop in the number of new post-hoc explainability methods for image classifiers was observed after the release of the Vision Transformer (ViT) \cite{dosovitskiyImage2020}, due to a partial shift in focus from developing model-agnostic methods, to exploring the self-attention maps of Vision Transformers for explainability \cite{caronEmerging2021,zhouRefiner2021}.
However, many newer Transformer-based image classifiers, such as \cite{tuMaxViT2022,liuSwin2021}, make ViT-specific explainability methods inapplicable, leaving only a few model-agnostic approaches like RISE \cite{petsiukRISE2018} to be applicable to these and any other classifier.

We can further categorize post-hoc XAI methods for image classifiers by their approach to producing explanation maps. Gradient-based methods, like Grad-CAM and Grad-CAM++ \cite{selvarajuGradCAM2017, chattopadhayGradCAM2018}, produce explanations using the model's gradients. Grad-CAM \cite{selvarajuGradCAM2017} produces explanations using a weighted sum of the feature maps of the final layer before classification.
The weights are computed via global average pooling of the gradient for each feature map with respect to the output class. These methods are simple and intuitive but face gradient-related issues, including noise and saturation from the activation functions \cite{adebayoSanity2018, pascanuDifficulty2013}. 
Additionally, because they utilize intermediate feature maps, they are not model-agnostic. Perturbation-based approaches observe how the model's outputs vary when the input is distorted. These methods can be model-agnostic, like RISE \cite{petsiukRISE2018}, LIME \cite{ribeiroWhy2016} and SHAP \cite{lundbergUnified2017}, or utilize intermediate feature maps, such as Score-CAM \cite{wangScoreCAM2020}, Opti-CAM \cite{zhangOptiCAM2024}, TAME \cite{ntrougkasTAME2022}, and T-TAME \cite{ntrougkasTTAME2024}. RISE \cite{petsiukRISE2018} generates random masks and uses them to mask the input image. The output confidence scores are used as weights in the weighted sum of the masks. LIME \cite{ribeiroWhy2016} trains a surrogate model based on the model's responses after perturbations of the input, but relies on superpixels for image explanations, being too computationally expensive to produce less coarse explanations. SHAP \cite{lundbergUnified2017} approximates Shapley values with SHAP scores, which represent the contribution of each input feature to the final decision of the classifier. Computing the exact Shapley values for complex neural network-based models is not feasible for high-dimensional data (e.g., images), and the approximated SHAP scores produce misleading information about relative feature importance \cite{huangFailings2024}, resulting in poor performance in XAI metrics measuring faithfulness.
Score-CAM \cite{wangScoreCAM2020} uses the DNN's final layer's feature maps, claiming that they represent better perturbations.
Opti-CAM \cite{zhangOptiCAM2024}, like Score-CAM, uses the final layer's feature maps, but trains a weight vector during inference to maximize the model's confidence. CAM \cite{zhouLearning2016} is a purely response-based method, using only the final layer's feature maps and the global average pooling layer's output to produce explanation maps, which constrains its application to very specific architectures. SISE \cite{sattarzadehExplaining2021} and Ada-SISE \cite{sudhakarAdaSise2021} blur the boundaries between the categories of gradient-, perturbation-, and response-based methods by using the gradients of the model's predictions, combining intermediate feature maps, and using them to perturb the input.
TAME \cite{ntrougkasTAME2022} and T-TAME \cite{ntrougkasTTAME2024} (the latter being applicable not only to convolutional neural networks (CNNs), as TAME is, but also to Vision Transformer-based architectures) probe the model during training, utilizing feature maps from multiple layers and learning weights to combine them.
During inference, they produce explanations without perturbations, lowering computational requirements.
Therefore, they are trainable response-based approaches; however, their attention mechanism is trained prior to inference (in contrast to Opti-CAM).

The proposed P-TAME is a trainable perturbation-based approach, and unlike T-TAME (and TAME), it is also model-agnostic, imposing no constraints on the backbone architecture. Additionally, while P-TAME needs to be trained separately for each backbone to be explained, training is performed with a standardised procedure that is identical regardless of the backbone.
P-TAME is performant during inference, requiring only a single forward pass to produce explanations, and is easily trainable and applicable to any DNN-based image classifier. We note that the explanation maps produced by P-TAME are not related to the attention maps produced by Visual Transformers \cite{caronEmerging2021}, as P-TAME is specifically trained to produce explanation maps that highlight regions which are important for the backbone's decision, while ViT attention maps are a byproduct of the multi-head attention, whose value as explanations is unclear \cite{cheferTransformer2021}.
\section{P-TAME}
\label{sec:method}

\subsection{Method overview}
The process of yielding explanations for the predictions of image classifiers with P-TAME involves two main steps. 
The first step is to train an attention mechanism that generates explanation maps from feature maps.
In contrast to T-TAME, feature maps are never directly extracted from the backbone
network (the DNN whose decisions should be explained); instead, they are produced by an auxiliary classifier (whose weights are also frozen). 
Thus, the P-TAME method is model-agnostic: only the input images and the backbone's output predictions are required.
The second step involves using the trained attention mechanism to directly produce class-specific explanations for the backbone's predictions.

The pipeline of the proposed framework is illustrated side-by-side with the T-TAME pipeline in Fig.~\ref{fig:pipeline}, highlighting the main difference between the two methods, which is the introduction of the auxiliary classifier in P-TAME.

\subsection{Definitions}
Consider an image classifier network (a.k.a. backbone) $f \colon \mathcal{X} \rightarrow \mathbb{R}^C$ that maps an input image $x \in \mathcal{X}$ to a vector of logits $y = (y)_x = f(x) \in \mathbb{R}^C$, where $\mathcal{X}$ is the space of images and $C$ is the number of classes.
We denote the $c$-th element of $y$ as $y_c$. Let $c^* = \arg \max y$ be the model-truth class, i.e., the model's prediction, which can be contrasted with a ground-truth class provided by a labeled dataset.
Additionally, consider an auxiliary image classifier network $f_\text{aux} \colon \mathcal{X} \rightarrow \mathbb{R}^C$. The auxiliary classifier $f_\text{aux}$ is constrained to only CNN-based architectures, because they produce three-dimensional feature maps. 
We denote the feature map extracted from layer $l$ of the auxiliary classifier as $F_l \in \mathbb{R}^{d_l \times w_l \times h_l}$. 
Here, $d_l$, $w_l$, and $h_l$ are the number of channels, height, and width of the feature map, respectively. 
The attention mechanism of P-TAME takes as input multiple feature maps from different layers of the auxiliary classifier, to improve the resolution of the produced explanation maps, based on the findings of \cite{ntrougkasTTAME2024}.
Let $\mathcal{A}({F_{L}}) = E$ be the attention mechanism, where ${F_{L}}$ is the set of feature maps extracted from ${L} = \left\{ l_1, l_2, \dots, l_s \right\}$ layers, and $E \in \left[ 0, 1 \right]^{C \times w_E \times h_E}$ the class-specific explanation maps. Finally, we denote by $R = w_E \cdot h_E$ the resolution of the explanation maps.
\begin{figure}[t]
  \centering
  \begin{subfigure}{1\linewidth}
    \includegraphics[width=1\linewidth]{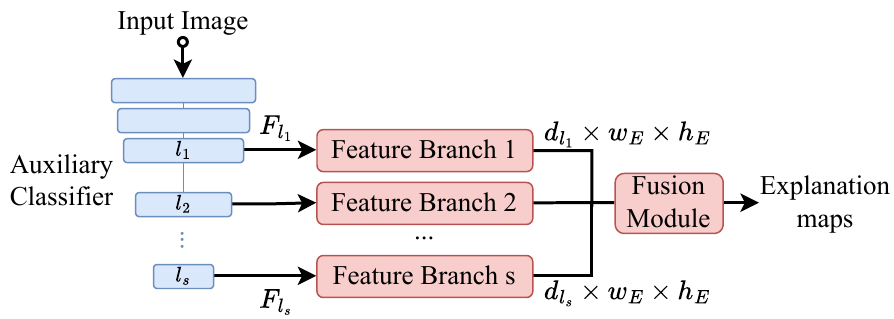}
    \caption{}
    \label{fig:attention-a}
  \end{subfigure}
  \begin{subfigure}{0.88\linewidth}
    \includegraphics[width=1\linewidth]{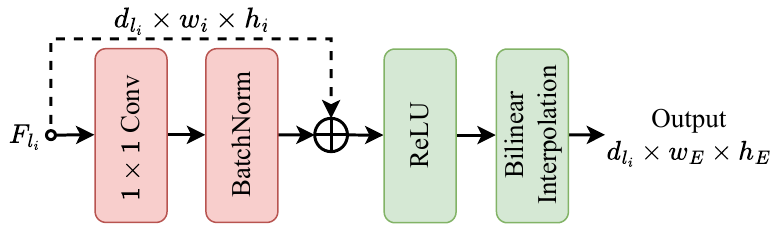}
    \caption{}
    \label{fig:attention-b}
    \vspace*{2mm}
  \end{subfigure}
  \begin{subfigure}{0.6\linewidth}
    \includegraphics[width=1\linewidth]{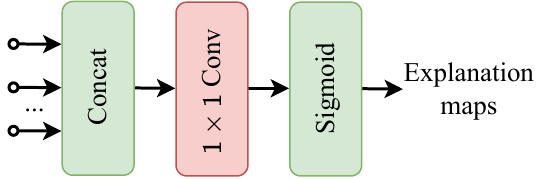}
    \caption{}
    \label{fig:attention-c}
  \end{subfigure}
  \caption{Structure of the attention mechanism of P-TAME (also used in T-TAME, though with different input). (a) Overview of the auxiliary classifier and the attention mechanism, (b) detailed structure of a feature branch of the attention mechanism, (c) detailed structure of the fusion module of the attention mechanism. The same color coding as in Fig.~\ref{fig:pipeline} is used to denote frozen / non-trainable / trainable components.}
  \label{fig:attention}
\end{figure}

\subsection{Auxiliary classifier and attention mechanism}
\label{ssec:aux}
The auxiliary classifier, a CNN pretrained on the same dataset as the backbone (e.g. ResNet-18 \cite{heDeep2016}, see Section~\ref{ssec: ex-setup} for experimentation details), extracts features that follow a predictable pattern: deeper layers capture semantically rich features, while earlier layers detect simple patterns or edges \cite{linFeature2017}. These features are three-dimensional, spatially consistent with the input image, and straightforward to process.
The P-TAME attention mechanism combines feature maps from various layers of the auxiliary classifier, which differ in channel count and spatial resolution.
Using these feature maps, it generates explanations that highlight the most salient input regions according to the backbone.
This adaptation involves processing each feature map individually and combining them to produce class-specific explanation maps, as illustrated in Fig.~\ref{fig:attention-a}.
We note that the auxiliary classifier’s role is to produce well-structured and diverse input-specific feature maps for the attention mechanism to combine, not to mirror the backbone’s logic. The trained P-TAME attention mechanism is responsible for reflecting the backbone's reasoning.

Feature maps ${F_{L}}$ from different layers of the auxiliary classifier are processed individually through a feature branch comprising a $1\times1$ convolution layer, batch normalization, a skip connection, an activation function, and bilinear interpolation (Fig.~\ref{fig:attention-b}).
Bilinear interpolation upscales smaller feature maps to match the resolution of the largest feature map. While feature maps extracted from deeper layers typically have lower resolutions, some architectures produce feature maps of equal resolution (e.g., architectures using inverted residual blocks), making bilinear interpolation necessary only when resolutions differ. 
All feature maps are scaled to the largest resolution, matching the resolution of the final explanation maps ($R$). The processed feature maps are concatenated and passed through a $1\times1$ convolution layer and a sigmoid activation function, producing class-specific explanation maps (Fig.~\ref{fig:attention-c}).
The sigmoid activation ensures that the resulting explanation maps have values in $\left[ 0, 1 \right]$.

\subsection{Training regime}
The attention mechanism we defined has to be trained to correctly combine the input feature maps into meaningful class-specific explanation maps. The auxiliary classifier's weights are frozen, so only the attention mechanism's weights need to be trained. This is done in a self-supervised manner, similarly to T-TAME (Fig.~\ref{fig:pipeline}). Specifically, images from the dataset used to train the backbone $f$ are input to both $f$ and the components of P-TAME: the auxiliary classifier $f_\text{aux}$ and the attention mechanism $\mathcal{A}$. During training, to measure how salient the explanations produced by P-TAME for the training image $x$ are, we first select the explanation $E_{c^*}$ (corresponding to the model truth class $c^*$) and use it to mask the image:
\begin{align}
    x_m = x \odot \text{up}_\text{bilinear}(E_{c^*}),
\end{align}
where $\text{up}_\text{bilinear}(\cdot)$ refers to bilinear interpolation, and $\odot$ refers to the Hadamard product. Then, the masked image is input a second time to the backbone to produce new predictions $f(x_m) = (y)_{x_m}$.
The masking procedure removes features that should be of low relevance to the classifier's prediction. 
After removing these features, we expect the confidence in the prediction to rise, as this is the basic premise of visual attention \cite{itti_model_1998}.
We measure the fidelity of the explanations through the response of the model with the cross-entropy loss: 
\begin{align}
    \mathcal{L}_\text{CE}(c^*, (y)_{x_m}) = - \log((y_{c^*})_{x_m}),
\end{align}
For this, we use the model-truth class $c^*$ instead of the classifier's original prediction $(y_{c^*})_x$, due to the training instability of soft cross-entropy \cite{caronEmerging2021}. 
The use of model-truth resembles many knowledge distillation methods (e.g., \cite{caronEmerging2021, chengExplaining2020}), and we can view the training of P-TAME as a form of distillation of the backbone model's reasoning.

With a naive minimization of the above loss, the all-ones mask $x_m = x \odot \mathbf{1} = x$ would be the trivial solution. To avoid this, we add a second loss term $\mathcal{L}_\text{Area}$, which penalizes the produced explanations based on how activated they are. For calculating this loss term, we consider not only the explanation for class $c^*$ but also for other classes, specifically for a uniformly sampled subset $S$ of $\{0, ..., C - 1\}$ with $c^* \in S$ and \(|S| = \lambda_{\text{rand}}\), where \(\lambda_{\text{rand}}\) is a hyperparameter. We use this subset $S$ instead of all $C$ classes to avoid excessive calculations. 
Thus, we define $\mathcal{L}_\text{Area} = \frac{1}{|E_{S}|} \sum E_{S}^{\lambda_\text{area}}$, where $\lambda_\text{area}$ is a hyperparameter and $|E_{S}|=|S|\cdot R$ is the total number of elements in the explanation maps \(E_{S}\).

Additionally, we want to encourage simpler explanations.
To minimize explanation complexity, we penalize the spatial variation within each explanation map belonging to the same subset $S$: 
\begin{align}
\mathcal{L}_\text{Variation} = \frac{1}{|E_{S}|} \sum_{c \in S} 
                               \left( \|\nabla_j E_{c,j,k}\|^2 +
                                     \|\nabla_k E_{c,j,k}\|^2 \right),
\end{align}
where \(\nabla_j E_{c,j,k} = E_{c,j+1,k} - E_{c,j,k}\) and \(\nabla_k E_{c,j,k} = E_{c,j,k+1} - E_{c,j,k}\) represent the spatial derivatives of \(E_c\).
Finally, the loss function used to train P-TAME is:
\begin{align}
  \mathcal{L}(c^*, (y)_{x_m}, E_{S}) = &\lambda_1 \mathcal{L}_\text{CE}(c^*, (y)_{x_m}) \nonumber \\
                               + &\lambda_2 \mathcal{L}_\text{Area}(E_{S}) \nonumber \\
                               + &\lambda_3 \mathcal{L}_\text{Variation}(E_{S}), \label{eq:loss}
\end{align}
where $\lambda_{\{1, 2, 3\}}$ are hyperparameters.
Here we can observe that P-TAME is wholly agnostic to the specific architecture of the image classifier $f$ that is being explained.

\subsection{Inference}
During inference, only one forward pass is required to compute explanation maps, as illustrated in Fig.~\ref{fig:pipeline}. The image is input to the backbone classifier to generate a prediction and to the auxiliary classifier to extract feature maps. Then, the feature maps are processed by the trained attention mechanism to generate class-specific explanation maps.
\section{Experiments}
\label{sec:results}

\begin{table*}[htp]\centering
  \centering
  \caption{Comparison of P-TAME with SoA methods using the AD, IC, MoRF and LeRF measures.}
  \label{tab:quantitative}
  \scriptsize
  \begin{booktabs}{
      column{3} = {r},
      column{4} = {r},
      column{5} = {r},
      column{6} = {r},
      column{7} = {r},
      column{8} = {r},
      column{9} = {r},
      column{10} = {r},
      column{11} = {r},
      column{12} = {r},
      cell{1}{3} = {c=3}{c},
      cell{1}{6} = {c=3}{c},
      cell{1}{9} = {c=2}{c},
      cell{3}{1} = {r=7}{l},
      cell{11}{1} = {r=7}{l},
      cell{19}{1} = {r=7}{l},
    }
    \toprule
              &                                           & AD↓             &                  &                  & IC↑              &                  &                  & ROAD (AUC)       &                  &               &             \\
    \cmidrule[lr]{3-5}\cmidrule[lr]{6-8}\cmidrule[lr]{9-10}
    Backbone  & Method                                    & 100\%           & 50\%             & 15\%             & 100\%            & 50\%             & 15\%             & MoRF↓            & LeRF↑            & R↑            & Fwd Passes↓ \\
    \midrule
    {VGG-16 \\ (acc@1: 71.59\% \cite{torchvision2016}) }  & Grad-CAM \cite{selvarajuGradCAM2017}      & 32.12\%         & 58.65\%          & 84.15\%          & 22.10\%          & 9.50\%           & 2.20\%           & 21.34\%          & 65.76\%          & 49            & \textbf{1}  \\
              & Grad-CAM++ \cite{chattopadhayGradCAM2018} & 30.75\%         & 54.11\%          & 82.72\%          & 22.05\%          & 11.15\%          & 3.15\%           & 22.57\%          & 64.54\%          & 49            & \textbf{1}  \\
              & RISE \cite{petsiukRISE2018}               & 8.74\%          & 42.42\%          & 78.70\%          & \uline{51.30\%}  & 17.55\%          & 4.45\%           & 22.72\%          & \textbf{69.25\%} & 49            & 4000        \\
              & Score-CAM \cite{wangScoreCAM2020}         & 27.75\%         & 45.60\%          & \uline{75.70\%}  & 22.80\%          & 14.10\%          & 4.30\%           & 22.12\%          & 66.66\%          & 49            & 512         \\
              & Ablation-CAM \cite{desaiAblationCAM2020}  & 34.87\%         & 49.23\%          & 76.96\%          & 19.25\%          & 11.45\%          & 3.65\%           & \uline{20.69\%}  & 66.95\%          & 49            & 2048        \\
              & Opti-CAM \cite{zhangOptiCAM2024}          & \textbf{2.23\%} & 42.66\%          & 87.97\%          & \textbf{85.91\%} & 20.78\%          & 2.18\%           & 26.24\%          & 61.21\%          & 49            & \uline{50}  \\
              & T-TAME \cite{ntrougkasTTAME2024}          & 9.33\%          & \uline{36.50\%}  & \textbf{73.29\%} & 50.00\%          & \textbf{22.45\%} & \textbf{5.60\%}  & \textbf{18.55\%} & 66.93\%          & \uline{784}   & \textbf{1}  \\
              & \textbf{P-TAME}                           & \uline{7.11\%}  & \textbf{33.39\%} & 76.06\%          & 49.06\%          & \uline{22.17\%}  & \uline{4.76\%}   & 24.78\%          & \uline{68.34\%}  & \textbf{3136} & \textbf{1}  \\
    \midrule
    {ResNet-50 \\ (acc@1: 76.13\% \cite{torchvision2016})} & Grad-CAM \cite{selvarajuGradCAM2017}      & 13.61\%         & 29.28\%          & 78.61\%          & 38.10\%          & 23.05\%          & 3.40\%           & 24.80\%          & \uline{73.38\%}  & 49            & \textbf{1}  \\
              & Grad-CAM++ \cite{chattopadhayGradCAM2018} & 13.63\%         & 30.37\%          & 79.58\%          & 37.95\%          & 23.45\%          & 3.40\%           & 25.95\%          & 72.34\%          & 49            & \textbf{1}  \\
              & RISE \cite{petsiukRISE2018}               & 11.12\%         & 36.31\%          & 82.05\%          & 46.15\%          & 21.55\%          & 3.20\%           & \textbf{23.42\%} & \textbf{73.74\%} & 49            & 8000        \\
              & Score-CAM \cite{wangScoreCAM2020}         & 11.01\%         & \textbf{26.80\%} & 78.72\%          & 39.55\%          & 24.75\%          & 3.60\%           & 27.01\%          & 72.10\%          & 49            & 512         \\
              & Ablation-CAM \cite{desaiAblationCAM2020}  & 13.58\%         & 30.33\%          & 79.62\%          & 37.05\%          & 22.30\%          & 3.50\%           & 25.78\%          & 72.23\%          & 49            & 8192        \\
              & Opti-CAM \cite{zhangOptiCAM2024}          & \textbf{1.27\%} & 38.49\%          & 90.00\%          & \textbf{90.87\%} & 24.60\%          & 1.79\%           & 32.83\%          & 62.97\%          & 49            & \uline{50}  \\
              & T-TAME \cite{ntrougkasTTAME2024}          & \uline{7.81\%}  & \uline{27.88\%}  & \uline{78.58\%}  & \uline{54.00\%}  & \textbf{27.50\%} & \textbf{4.90\%}  & \uline{24.61\%}  & 68.89\%          & \uline{784}   & \textbf{1}  \\
              & \textbf{P-TAME}                           & 8.35\%          & 28.95\%          & \textbf{77.53\%} & 50.00\%          & \uline{24.85\%}  & \uline{4.81\%}   & 26.13\%          & 71.27\%          & \textbf{3136} & \textbf{1}  \\
    \midrule
    {ViT-B-16 \\ (acc@1: 81.07\% \cite{torchvision2016})} & Grad-CAM \cite{selvarajuGradCAM2017}      & 37.19\%         & 40.74\%          & 73.11\%          & 12.75\%          & 12.30\%          & 5.40\%           & \uline{27.65\%}  & 71.92\%          & \uline{196}   & \textbf{1}  \\
              & Grad-CAM++ \cite{chattopadhayGradCAM2018} & 57.21\%         & 72.77\%          & 92.51\%          & 5.55\%           & 4.85\%           & 0.80\%           & 46.98\%          & 64.35\%          & \uline{196}   & \textbf{1}  \\
              & RISE \cite{petsiukRISE2018}               & 38.09\%         & 44.20\%          & 77.50\%          & 15.35\%          & 14.50\%          & 4.85\%           & 36.85\%          & \textbf{76.28\%} & 49            & 8000        \\
              & Score-CAM \cite{wangScoreCAM2020}         & 35.50\%         & 42.16\%          & 80.86\%          & 8.90\%           & 10.55\%          & 2.95\%           & 32.25\%          & 62.65\%          & \uline{196}   & 768         \\
              & Ablation-CAM \cite{desaiAblationCAM2020}  & 38.09\%         & 44.20\%          & 77.50\%          & 15.35\%          & 14.50\%          & 4.85\%           & 33.30\%          & 72.27\%          & \uline{196}   & 768         \\
              & Opti-CAM \cite{zhangOptiCAM2024}          & \textbf{0.15\%} & 67.29\%          & 93.36\%          & \textbf{98.07\%} & 13.29\%          & 1.88\%           & 47.62\%          & 54.51\%          & \uline{196}   & \uline{50}  \\
              & T-TAME \cite{ntrougkasTTAME2024}          & 8.19\%          & \uline{23.64\%}  & \uline{72.89\%}  & 38.35\%          & \uline{40.40\%}  & \uline{9.40\%}   & \textbf{24.66\%} & \uline{74.97\%}  & \uline{196}   & \textbf{1}  \\
              & \textbf{P-TAME}                           & \uline{7.50\%}  & \textbf{19.63\%} & \textbf{62.69\%} & \uline{47.47\%}  & \textbf{43.45\%} & \textbf{11.86\%} & 33.89\%          & 73.01\%          & \textbf{3136} & \textbf{1}  \\
    \bottomrule
  \end{booktabs}
\end{table*}

\begin{figure*}[t]
    \centering
    \includegraphics[width=0.92\linewidth]{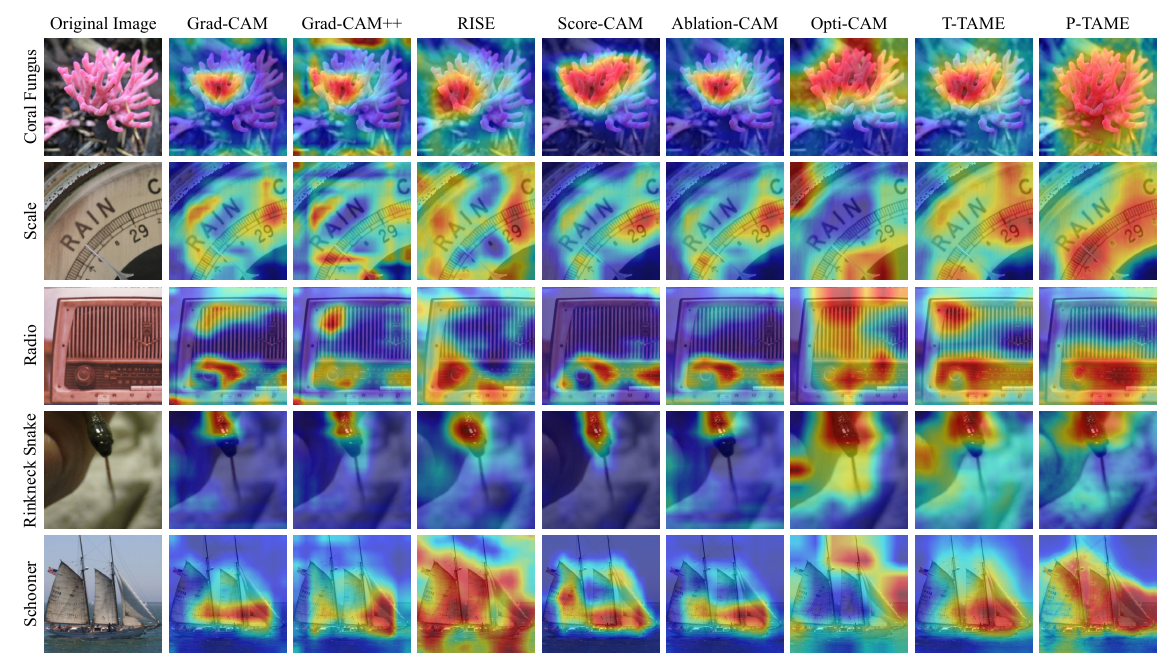}
    \caption{Explanation maps produced by different methods for the ResNet-50 \cite{heDeep2016} backbone. The model truth class is shown on the left.}
    \label{fig:qualitative}
\end{figure*}
\subsection{Experimental setup}
\label{ssec: ex-setup}
We perform a comprehensive evaluation of P-TAME by comparing it both quantitatively and qualitatively against SoA explainability methods across 3 backbone image classifiers: VGG-16 \cite{simonyanVery2015}, ResNet-50 \cite{heDeep2016}, and ViT-B-16 \cite{dosovitskiyImage2020}. 
For measuring explanation quality, we adopt evaluation measures that are widely used in the domain. 
We also report the resolution of the produced explanation maps before rescaling and measure the computational requirements of different explainability methods by reporting the number of forward passes required to produce an explanation.
Additionally, we perform a sanity check, per \cite{adebayoSanity2018}, to determine that the explanations produced by P-TAME are sensitive to the parameters of the backbone model.
Furthermore, we perform an ablation study examining the effects of different choices of auxiliary classifiers both quantitatively and qualitatively. 
In the latter ablation, we compare three lightweight image classifiers: ResNet-18 \cite{heDeep2016}, MobileNetV3 \cite{howardSearching2019}, and MnasNet \cite{tanMnasNet2019}. 
Besides assessing differences in the explanation quality, we also compare the computation requirements imposed by each auxiliary classifier (measured in GFLOPs) and quantify how the features extracted from each different layer of the auxiliary classifiers contribute to the final explanation maps.

\textbf{Dataset:} We use the ImageNet ILSVRC 2012 dataset \cite{krizhevskyImageNet2012}. The training subset of it (1,281,167 images) is used for training P-TAME, while two subsets of 2000 images each from the ILSVRC 2012 evaluation set are used as our validation and testing sets, respectively. The number of image classes (in this dataset and the pre-trained backbones and auxiliary classifiers used in our experiments) is $C=1000$.

\textbf{Models:} For the backbone image classifiers VGG-16 \cite{simonyanVery2015}, ResNet-50 \cite{heDeep2016} and ViT-B-16 \cite{dosovitskiyImage2020}, we use their ImageNet-pretrained instances available in the \verb|torchvision| library \cite{torchvision2016}.
These classifiers represent three very distinct evolutionary phases in the field of DNN-based image classification, each introducing significant architectural shifts w.r.t. their predecessors, i.e., the 2-dimensional convolution layer, the skip connection, and the multi-head attention layer. 
A ResNet-18 \cite{heDeep2016} model, also pretrained on ImageNet and retrieved from \verb|torchvision|, is used as our auxiliary classifier, chosen because it strikes a good balance between performance and computational requirements. Feature maps are extracted from the outputs of the last four residual blocks of ResNet-18. Other choices of auxiliary classifiers are considered in the ablation study (Section~\ref{ssec:abl}). 

\textbf{Training:} We train P-TAME's attention mechanism on the ImageNet dataset for one epoch, using a batch size of $64$ images, the largest batch size that our GPU can support (following \cite{smithDisciplined2018}). 
In our experiments, we observed that further training did not improve the performance of the method. 
We use the AdamW optimizer \cite{loshchilovDecoupled2018} and the OneCycleLR learning rate scheduler \cite{smithSuperConvergence2019}, setting the maximum learning rate to either $10^{-4}$ or $10^{-3}$. The hyperparameter $\lambda_\text{rand}$ is set equal to the batch size.
Prior to this training, to determine appropriate values for the hyperparameters introduced in the loss function (Eq.~\ref{eq:loss}), we utilize Bayesian optimization, specifically the BoTorch framework \cite{balandatBoTorch2020}. 
Bayesian optimization is a well-established technique for serial optimization of costly-to-evaluate black-box functions, such as the training and evaluation of a neural network. 
Bayesian optimization involves conducting several trials, and for each one, we train for a single epoch and then evaluate it using the MoRF and LeRF measures (see ``Evaluation measures'', below) computed on the validation set. 
The search space is greatly compacted by constraining the loss term weights by $\sum_i \lambda_i = 1$, allowing $\lambda_1$ and $\lambda_2$ to take values in the range $[0,1]$ with the condition that $\lambda_1 + \lambda_2 < 1$ and setting $\lambda_3 = 1 - \lambda_1 - \lambda_2$. 
Also, $\lambda_\text{area}$ is allowed to take a value from set $\{0.5, 1, 2\}$. With only 5 initial random trials and 15 subsequent trials of Bayesian optimization (i.e., 20 trials in total), the hyperparameter optimization procedure converges. The exact parameters to reproduce the reported results are included in the released source code.

\textbf{Evaluation measures:} For evaluating explainability methods for image classifiers, the most crucial aspect of explanations that we want to quantify is their ``faithfulness'', or how much they align with the image classifier that is being explained. 
The approach most frequently used in the domain is to perturb the input image, using the explanation map as a mask of the image to observe how the confidence in the original prediction changes. 
We use 8 measures to capture ``faithfulness''. The most widely employed measures, Average Drop (AD) and Increase in Confidence (IC), are defined as \cite{chattopadhayGradCAM2018}:
\begin{align}
    \mathrm{AD}(v) &= \sum_{x}\frac{\max\lbrace 0, (y_{c^*})_x - (y_{c^*})_{x_{m(v)}}\rbrace}{
        \Upsilon},\\
    \mathrm{IC}(v) &= \sum_{x}\frac{\mathrm{int}\left((y_{c^*})_{x_{m(v)}}>(y_{c^*})_x\right)}{\Upsilon},
\end{align}
where $\Upsilon$ represents the number of test images. Here, $x_{m(v)}$ is the masked image, with a threshold applied to the mask to select the top $v\%$ highest-valued pixels of the explanation map $E_{c^*}$.
We also use the MoRF and LeRF measures \cite{rongConsistent2022}:
\begin{align}
    \mathrm{MoRF}(v) &= \sum_{x}\frac{\mathbb{I}((c^*)_{x_{\hat{m}(v)}} = (c^*)_x)}{\Upsilon}, \\
    \mathrm{LeRF}(v) &= \sum_{x}\frac{\mathbb{I}((c^*)_{x_{\check{m}(v)}} = (c^*)_x)}{\Upsilon},
\end{align}
where $\mathbb{I}()$ is an indicator function that returns 1 if the condition is true and 0 otherwise, $x_{\hat{m}(v)}$ and $x_{\check{m}(v)}$ denote the image masked with a binary mask which selects the top $v\%$ highest or lowest valued pixels of the explanation map, respectively.
The masking procedure is a type of image infilling, described in \cite{rongConsistent2022}.
We threshold the mask at percentages  $(10\%, 20\%, 30\%, 40\%, 50\%, 70\%, 90\%)$ as in \cite{rongConsistent2022} to assess the effectiveness of the explanation in ranking pixel importance.
The area under the curve of the resulting accuracies is computed to aggregate the results from the various thresholds. 
A low MoRF indicates that the explanation map correctly identifies the most significant image regions for the prediction, while a high LeRF signifies accurate identification of the least significant regions. MoRF and LeRF are independent of mask distribution and rely solely on pixel ranking, with the infilling procedure mitigating input distribution shifts, which particularly impact CNNs\cite{madalaCNNs2023}.

\begin{figure}[t]
    \centering
    \includegraphics[width=0.75\linewidth]{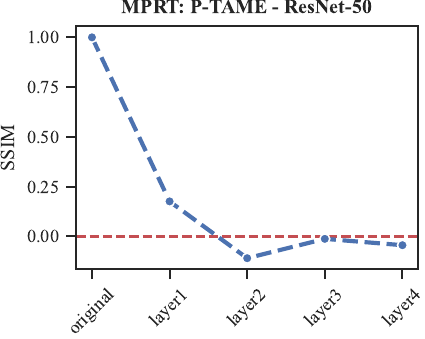}
    \caption{Model Parameter Randomization Test (MPRT) for P-TAME on the ResNet-50 \cite{heDeep2016} backbone. We observe a sharp drop in SSIM as the backbone's layers (indicated on the horizontal axis) are randomized, showcasing that P-TAME passes the sanity check for saliency maps \cite{adebayoSanity2018}.}
    \label{fig:mprt}
\end{figure}

\subsection{Quantitative results and comparisons}
In Table~\ref{tab:quantitative}, our proposed P-TAME method is compared with the following SoA methods: Grad-CAM \cite{selvarajuGradCAM2017}, Grad-CAM++ \cite{chattopadhayGradCAM2018}, RISE \cite{petsiukRISE2018}, Score-CAM \cite{wangScoreCAM2020}, Ablation-CAM \cite{desaiAblationCAM2020}, and T-TAME \cite{ntrougkasTTAME2024}.
We selected these specific methods because they are among the most widely used and performant methods of their respective class (gradient-, perturbation-, and response-based approaches).
From the results, we observe that for the ViT-B-16 backbone, we obtain top performance in the AD and IC measures, except for the $v=100\%$ threshold, which is dominated by Opti-CAM across different backbones. 
However, Opti-CAM exhibits the worst performance in the more challenging AD(15\%), IC(15\%), and ROAD measures.
For the CNN models VGG-16 and ResNet-50, we obtain near-top scores for the AD and IC measures, competing in performance only with T-TAME and the model-agnostic perturbation method RISE. 
In the MoRF and LeRF measures, which signal if the ordering of pixels by importance is correct, P-TAME provides mixed results. 
This is mostly caused by the fact that the explanation maps produced by P-TAME have a much higher resolution, and providing a good ordering of $R$ pixels is much simpler for
lower resolutions. 
This is further elucidated in Section~\ref{ssec:qual}.
Still, the fact that P-TAME generates explanation maps in a single forward step and can be applied to any image classifier architecture is a significant advantage compared to more computationally intense methods (e.g., RISE) or more restrictive feature map extraction methods (e.g., Grad-CAM, T-TAME).

In Fig.~\ref{fig:mprt}, the outcome of a Model Parameter Randomization Test (MPRT), a sanity check to determine whether an XAI method is sensitive to the backbone's parameters \cite{adebayoSanity2018}, is shown for P-TAME applied to the ResNet-50 backbone. Structure Similarity Index Measure (SSIM) values are calculated between the explanations produced for the pretrained ResNet-50 model and the explanations produced after randomizing all its parameters up to the layer named in the x-axis, following a bottom-up approach as recommended in \cite{hedstromFresh2024}. An SSIM value near zero denotes no similarity between the compared explanation maps. We observe a sharp decrease in SSIM immediately, demonstrating that P-TAME is highly sensitive to the randomization of the backbone and thus passes the sanity check for saliency maps. 

\begin{table}[tbp]\centering
  \caption{Ablation study: different choices of auxiliary classifier.}
  \label{tab:ablation}
  \scriptsize
  \begin{booktabs}{
      column{1-6} = {r},
      row{1-2} = {l},
    }
    \toprule
    Auxiliary & ResNet- & MobileNet- & MnasNet & ResNet- & ResNet-\\
Classifier & 18 \cite{heDeep2016} & V3 \cite{howardSearching2019} & \cite{tanMnasNet2019} & 50 \cite{heDeep2016} & 18* \cite{heDeep2016}\\
    \midrule
    AD 100\%↓ & \uline{8.35\%} & 11.15\% & 14.11\% & \textbf{7.81\%} & 9.13\%\\
IC 100\%↑ & \uline{50.00\%} & 42.26\% & 38.29\% & \textbf{54.00\%} & 44.59\%\\
AD 50\%↓ & \uline{28.95\%} & 36.59\% & 43.24\% & \textbf{27.88\%} & 36.55\%\\
IC 50\%↑ & \uline{24.85\%} & 19.64\% & 16.22\% & \textbf{27.50\%} & 18.40\%\\
AD 15\%↓ & \textbf{77.53\%} & 79.81\% & 83.83\% & \uline{78.58\%} & 84.27\%\\
IC 15\%↑ & \uline{4.81\%} & 4.56\% & 2.68\% & \textbf{4.90\%} & 2.78\%\\
    \midrule 
MoRF↓ & 26.13\% & \textbf{24.28\%} & 26.77\% & \uline{24.61\%} & 28.32\%\\
LeRF↑ & \textbf{71.27\%} & \uline{69.13\%} & 66.76\% & 68.89\% & 64.53\%\\
    \midrule
R↑ & \textbf{3136} & 49 & 49 & \uline{784} & \textbf{3136}\\
GFLOPs↓ & 46.42 & \textbf{24.97} & \uline{26.89} & {104.89} & 46.42\\
    \midrule
Layer 1 & 6.73\% & 11.11\% & 3.67\% & - & 6.72\%\\
Layer 2 & 13.44\% & 11.10\% & 32.09\% & 14.31\% & 13.48\%\\
Layer 3 & 26.68\% & 11.06\% & 31.55\% & 28.61\% & 26.81\%\\
Layer 4 & 53.15\% & 66.72\% & 32.69\% & 57.08\% & 52.99\%\\
    \bottomrule
  \end{booktabs}
\end{table}

\subsection{Ablations}
\label{ssec:abl}
In Table~\ref{tab:ablation}, we examine different auxiliary classifiers for explaining the ResNet-50 backbone, initially comparing our choice of ResNet-18 with MobileNetV3 \cite{howardSearching2019} and MnasNet \cite{tanMnasNet2019} (again, models pretrained on ImageNet, retrieved from \cite{torchvision2016}). We observe that smaller auxiliary classifiers offer computational advantages but produce coarser explanation maps due to lower feature resolution.
This also results in modest improvements in MoRF and LeRF for MobileNetV3, as it is easier to produce explanation maps with $49 = 7^2$ elements than with $3136 = 56^2$ elements. Overall, however, using ResNet-18 outperforms using any of the other two models, indicating a clear tradeoff between compute and explanation quality in selecting the auxiliary classifier.
We also note that the contribution of feature maps extracted from different layers to the final explanation maps varies greatly across classifiers. These contributions, calculated by processing the fusion module's trained weights (Fig.~\ref{fig:attention-c}) and grouping them based on which feature branch they correspond to (Fig.~\ref{fig:attention-a}), show that the deeper layer's feature maps consistently contribute more. In ResNet-18, contributions increase steadily with deeper layers, while MobileNetV3 and MnasNet show near-equal contributions across layers.
This difference is due to the architectures of MobileNetV3 and MnasNet, which use strided convolutions followed by inverted residual blocks, in contrast to the typical residual blocks found in ResNet-18. Inverted residual blocks are computationally efficient but yield feature maps with fewer channels and small spatial dimensions, making it harder for P-TAME to transform these feature maps into class-specific explanation maps.

In the case that the auxiliary classifier is identical to the backbone being explained (i.e., in this case, ResNet-50), P-TAME degenerates to T-TAME (Fig.~\ref{fig:pipeline}; in this experiment, feature maps from 3 layers are used, following the T-TAME protocol \cite{ntrougkasTTAME2024}). Using the much larger ResNet-50 model as an auxiliary classifier yields moderate improvements across most measures, but more than doubles the computational requirements (GFLOPs). Note that T-TAME is model-specific; i.e., when the backbone also serves as the feature map generator (instead of using an auxiliary classifier), changes in the backbone being explained necessitate architectural changes in the attention mechanism of the explanation method. 

In a final ablation, we experiment with using as auxiliary classifier a ResNet-18 model trained on Imagenette \cite{Howard_Imagenette_2019}, a small subset of ImageNet containing just 10 classes and $1\%$ of the original images. This ablation examines the extent to which it is necessary to use an auxiliary classifier pretrained on the same dataset as the backbone being explained. The Imagenette-pretrained auxiliary classifier is denoted as ``ResNet-18*'' in Table~\ref{tab:ablation}. We observe from these results near-top performance (compared to using other auxiliary classifiers) for the AD and IC at higher thresholds (Table~\ref{tab:quantitative}), but a significant drop at the 15\% threshold and for the ROAD measures. This demonstrates that despite using an auxiliary classifier trained on a much smaller dataset, P-TAME continues to produce meaningful explanations; however, achieving SoA results requires a feature map generator trained on the backbone’s original training data.
\begin{figure}[t]
    \centering
    \includegraphics[width=0.91\linewidth]{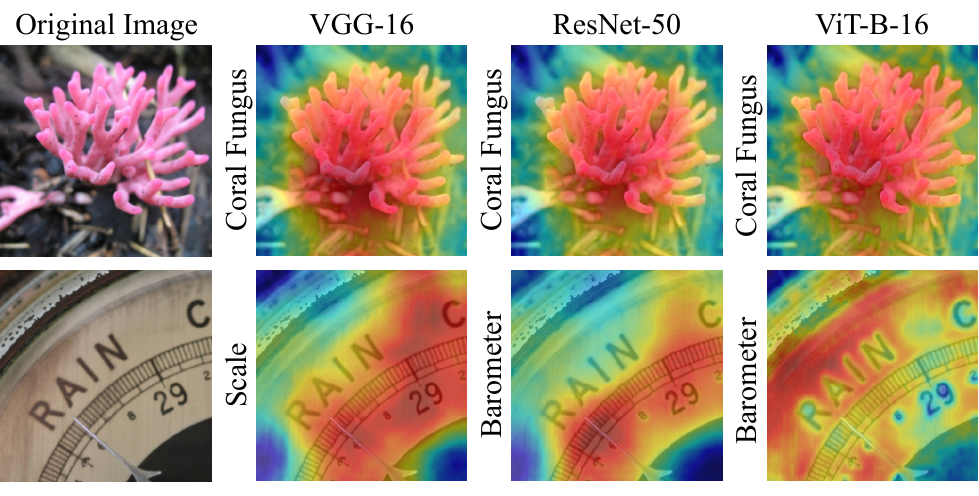}
    \caption{Explanation maps produced for the VGG-16 \cite{simonyanVery2015}, ResNet-50 \cite{heDeep2016}, and ViT-B-16 \cite{dosovitskiyImage2020} backbones. The model-truth class of the original image according to each backbone is shown on the left.}
    \label{fig:qualitative2}
\end{figure}

\subsection{Qualitative results}
\label{ssec:qual}
In Fig.~\ref{fig:qualitative}, explanation maps produced for the ResNet-50 backbone using P-TAME and the SoA methods of Table~\ref{tab:quantitative} are shown, following the findings of \cite{chowdhuryAre2024} on the importance of complementing quantitative evaluation with qualitative analysis. 
We select the ResNet-50 backbone for this qualitative comparison because it is one of the most widely used CNN architectures, and most of the compared explainability methods were developed for CNNs.
We observe that P-TAME produces the most activated explanation maps, followed by T-TAME and RISE. 
P-TAME correctly highlights the entire class when it can be localized (rows 1, 4, 5). 
In cases where the class cannot be localized, P-TAME correctly highlights salient features, in line with methods that directly make use of features extracted from the backbone. 
Along with the good quantitative results in Table~\ref{tab:quantitative}, this shows that P-TAME produces high-quality explanation maps in a single forward pass without requiring any backbone architecture-specific tailoring to extract and process feature maps.
The only other model-agnostic method, RISE, besides requiring 8000 forward passes to produce the shown explanation maps, produces noisier results, especially in cases where the class is not easily localizable (rows 2, 3, 5).

In Fig.~\ref{fig:qualitative2}, we compare explanation maps produced for our three backbones (VGG-16, ResNet-50, and ViT-B-16) using P-TAME.
For the first image, which shows a localizable class, the explanation maps are similar across backbones. 
However, for the second image, whose class cannot be easily localized to a specific region of the image, the ViT-B-16 backbone, the most performant model out of the three in terms of classification performance (see 1st column of Table~\ref{tab:quantitative}), shows the highest level of detail in its explanation. For example, the number ``29'' in the second image is shown to have low importance for the model-truth prediction.
For less performant models, like VGG-16, the explanations show much less detail, even though the resolution of the explanation map is the same as for ViT-B-16. This indicates a performance-explainability trade-off, i.e., that a higher-performing classifier can support the generation of more detailed explanations for it.

\begin{figure}[t]
    \centering
    \includegraphics[width=1\linewidth]{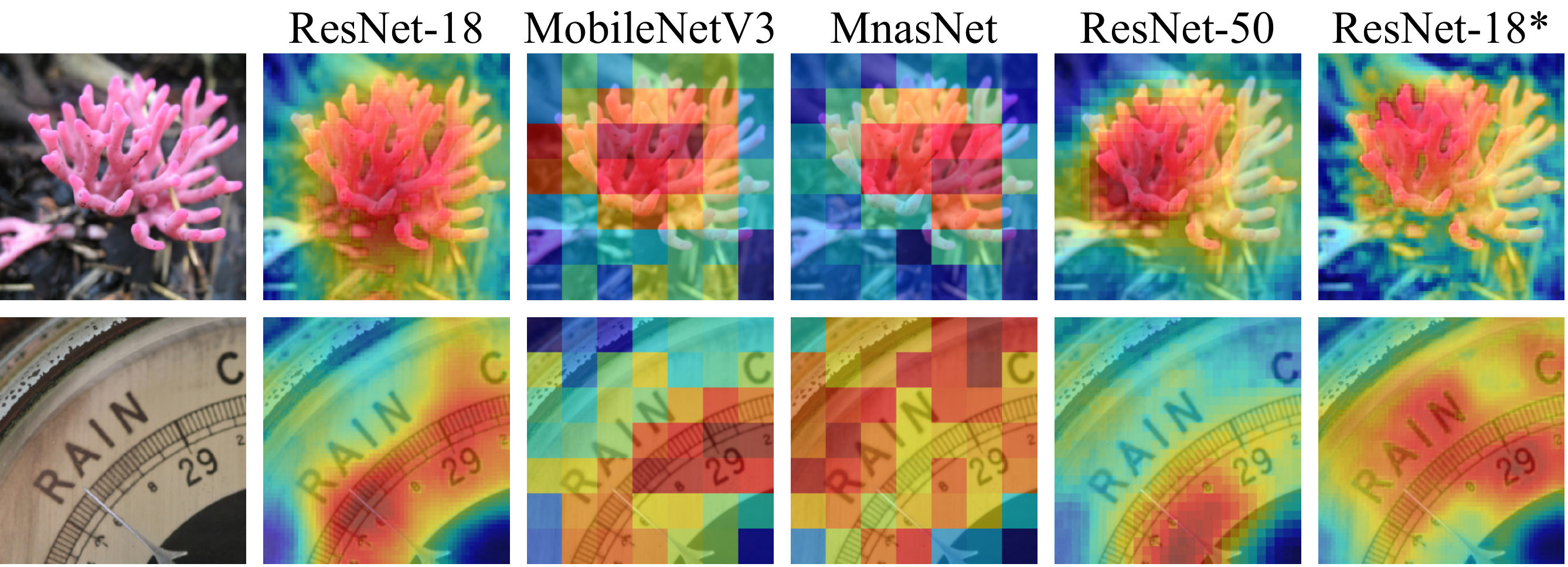}
    \caption{Explanation maps produced for the ResNet-50 \cite{heDeep2016} backbone, using different auxiliary classifiers. The model-truth class of these images is ``coral fungus'' and ``barometer'', respectively. To illustrate the raw resolution of the generated explanations, we upscale the explanation maps using nearest-neighbor interpolation (avoiding the smoothing effects of bilinear interpolation, used in other figures).}
    \label{fig:qualitative3}
\end{figure}

In Fig.~\ref{fig:qualitative3}, explanation maps produced for the ResNet-50 backbone using different auxiliary classifiers are shown. 
To demonstrate the differences in the resolution of the explanation maps, we use nearest-neighbor upscaling instead of bilinear interpolation used in other figures. 
These examples indicate that the different auxiliary classifiers generally agree on which are the most and least important parts of the image for the classification decision of the backbone, although the explanations produced by MobileNet-V3 and MnasNet are much coarser, due to their much lower resolution. 
This is in accord with the expected behavior since, in each case, the predictions of the same backbone are being explained. 
The explanation maps produced using the ResNet-50 auxiliary classifier are also coarser than those produced using ResNet-18, but more focused, explaining the moderate performance gains in Table~\ref{tab:ablation}. Finally, the explanation maps generated using the feature maps of ResNet-18 trained on Imagenette (denoted as ``ResNet-18*'') are noisier, with less well-ordered pixel importance, highlighting the importance of using a well-trained auxiliary classifier. 

\section{Conclusions}
This paper presented P-TAME, a method for explaining DNN image classifiers by training an attention mechanism to combine feature maps produced by an auxiliary classifier into explanation maps, highlighting the important regions for the backbone model's prediction. 
P-TAME improves upon the paradigm established by T-TAME, extending it by decoupling the input of the attention mechanism responsible for producing explanations from the intermediate feature maps of the backbone being explained. 
This makes P-TAME a model-agnostic method, rendering it much more widely applicable. 
P-TAME produces explanation maps in a single forward pass during inference, while producing explanations that are on par with or better than those of the SoA explainability approaches. 
An important current limitation of P-TAME is the need for an auxiliary classifier trained on the backbone’s original training data. A promising future direction is to investigate training the auxiliary classifier via knowledge distillation on the backbone itself, mitigating the need to procure the original training data and better tailoring it to the backbone being explained.
{
\small
\bibliographystyle{ieeenat_fullname}
\bibliography{main,bibliography}
}

\end{document}